\documentclass[runningheads]{llncs}
\usepackage[T1]{fontenc}
\usepackage{amsmath} 
\usepackage{subcaption}
\usepackage{placeins}

\usepackage{graphicx,verbatim}

\usepackage{hyperref}
\usepackage{color}

\usepackage{multirow}
\begin{document}
\title{MedLoRD: A Medical Low-Resource \\ Diffusion Model for High-Resolution \\ 3D CT Image Synthesis}

 \author{Marvin Seyfarth\inst{1,4}\orcidID{0009-0007-4213-0750} \and
 Salman Ul Hassan Dar\inst{1,2,3,4}\orcidID{0000-0002-7603-4245} \and
 Isabelle Ayx \inst{5}\orcidID{0000-0003-3507-9006} \and
 Matthias Alexander Fink\inst{6}\orcidID{0000-0002-0189-7070}\and
 Stefan O. Schoenberg\inst{2,5}\orcidID{0000-0002-3115-8367} \and
 Hans-Ulrich Kauczor\inst{6}\orcidID{0000-0002-6730-9462} \and
 Sandy Engelhardt\inst{1,2,4}\orcidID{0000-0001-8816-7654 }}
 \author{Marvin Seyfarth\inst{1,4}\and
 Salman Ul Hassan Dar\inst{1,2,3,4}\and
 Isabelle Ayx \inst{5} \and
 Matthias Alexander Fink\inst{6}\and
 Stefan O. Schoenberg\inst{2,5}\and
 Hans-Ulrich Kauczor\inst{6}\and
 Sandy Engelhardt\inst{1,2,4}}
 \authorrunning{M. Seyfarth et al.}
 \titlerunning{MedLoRD: High-Resolution 3D CT Synthesis}
 \institute{Institute for Artificial Intelligence in Cardiovascular Medicine, Department of Cardiology, Angiology, Pneumology, Heidelberg University Hospital, Heidelberg, Germany \\ \and
     AI Health Innovation Cluster (AIH), Germany \and
     Heidelberg Faculty of Medicine, Heidelberg University, Germany\and
     German Centre for Cardiovascular Research (DZHK), Partner site Heidelberg/Mannheim, Germany \and
     Department of Radiology and Nuclear Medicine, University Medical Center Mannheim, Germany \and 
     Clinic for Diagnostic and Interventional Radiology, Heidelberg University Hospital, Heidelberg, Germany 
     \\
     \email{Marvin.Seyfarth@med.uni-heidelberg.de}
 }
\maketitle           
\begin{abstract}
Advancements in AI for medical imaging offer significant potential. However, their applications are constrained by the limited availability of data and the reluctance of medical centers to share it due to patient privacy concerns. 
Generative models present a promising solution by creating synthetic data as a substitute for real patient data. However, medical images are typically high-dimensional, and current state-of-the-art methods are often impractical for computational resource-constrained healthcare environments. 
These models rely on data sub-sampling, raising doubts about their feasibility and real-world applicability. 
Furthermore, many of these models are evaluated on quantitative metrics that alone can be misleading in assessing the image quality and clinical meaningfulness of the generated images.
To address this, we introduce MedLoRD, a generative diffusion model designed for computational resource-constrained environments. 
MedLoRD is capable of generating high-dimensional medical volumes with resolutions up to 512$\times$512$\times$256, utilizing GPUs with only 24GB VRAM, which are commonly found in standard desktop workstations.
MedLoRD is evaluated across multiple CT imaging applications, including Coronary Computed Tomography Angiography and Lung Computed Tomography datasets.
Extensive evaluations through radiological evaluation, relative regional volume analysis, adherence to conditional masks, and downstream tasks show that MedLoRD generates high-fidelity images, surpassing the capabilities of current state-of-the-art generative models for medical image synthesis in computational resource-constrained environments.


\end{abstract}

\section{Introduction}

Generative AI has emerged as a promising solution for models to learn data distributions and generate synthetic samples that can expand datasets or serve as privacy-preserving substitutes for real patient data \cite{meddiff,brainsynth}.\\
Despite progress, challenges remain in real-world implementation. Medical imaging data is high-dimensional, making it difficult for generative models to operate efficiently with limited GPU resources. Some approaches downsample data to capture global structures \cite{peng2023,dorjsembe2023conditional} but risk losing fine-grained details critical for diagnosis, while others apply post-processing techniques \cite{maisi} that may introduce artifacts.
Additionally, many studies rely on high-end GPUs that are often unavailable in clinical environments, creating a gap between research and practical implementation \cite{maisi,medsyn}.  
\\
To address these challenges, we introduce MedLoRD, a Medical Low Resource latent Diffusion model for medical image synthesis. MedLoRD uses Vector Quantized Variational Autoencoder Generative Adversial Networks (VQ-VAE GANs) for encoding and a 3D U-Net for image denoising in the latent space. Leveraging a streamlined VQ-VAE design, memory-efficient attention mechanisms, and a lightweight conditioning module, MedLoRD can be trained effectively on small datasets and synthesizes high-fidelity 3D volumes up to 512×512×256 in resolution on 24GB GPUs, without requiring post-processing. To assess the model’s clinical relevance, we conduct extensive evaluations, including radiological assessment, regional volume analysis, quantitative metrics such as Fréchet Inception Distance (FID) and DICE similarity score (DSC), as well as performance testing on downstream tasks. Overall, we propose a lightweight generative model that produces high-resolution 3D images under realistic computational constraints, making it well-suited for real-world deployment in medical imaging settings.

\section{Methodology}
MedLoRD builds on the latent diffusion framework \cite{LDM}, incorporating a set of targeted modifications aimed at efficient, high-fidelity 3D medical image synthesis (see Fig.~\ref{fig:medlord_architecture}). A VQ-VAE GAN serves as the first-stage compression network, projecting volumetric data into a discrete latent space. We use a large codebook of 16,384 embeddings with a feature dimension of 8, which allows for a compact autoencoder architecture, featuring fewer convolutional channels and reduced depth, without sacrificing reconstruction quality.
To address the high computational cost of distance calculations in large-codebook VQ-VAEs, we implement a batched distance computation strategy. This reduces memory usage and accelerates training for large 3D volumes without loss of precision.
Latent diffusion is performed by progressively adding noise over T timesteps, and training a U-Net to predict a combination of noise and the original latent sample \cite{LDM}. We replace all 2D convolutions with 3D convolutions and enable Xformers’ flash attention in the U-Net’s self-attention layers, reducing memory complexity to O(1)\cite{flash_attention}.
For conditional generation, we propose a ControlNet-inspired approach \cite{controlnetxs}. Instead of duplicating the encoder and middle blocks of the pretrained U-Net, we initialize a lightweight encoder, with 50\% fewer channels, and train it from scratch. This encoder is trained to modulate features of the pretrained backbone, effectively integrating conditioning inputs while retaining the generative power of the base model. This design balances memory efficiency with strong conditioning fidelity. 
Our models were implemented using the MONAI library \cite{monai} with inspiration from  \footnote{\url{https://github.com/Warvito/monai-vqvae-diffusion/tree/main}} and\footnote{\url{https://github.com/Project-MONAI/GenerativeModels}}. The implementation, along with configuration files, are publicly available at: \url{https://github.com/Cardio-AI/medlord}
\begin{figure}[t]  
    \centering
    \includegraphics[width=0.85\textwidth]{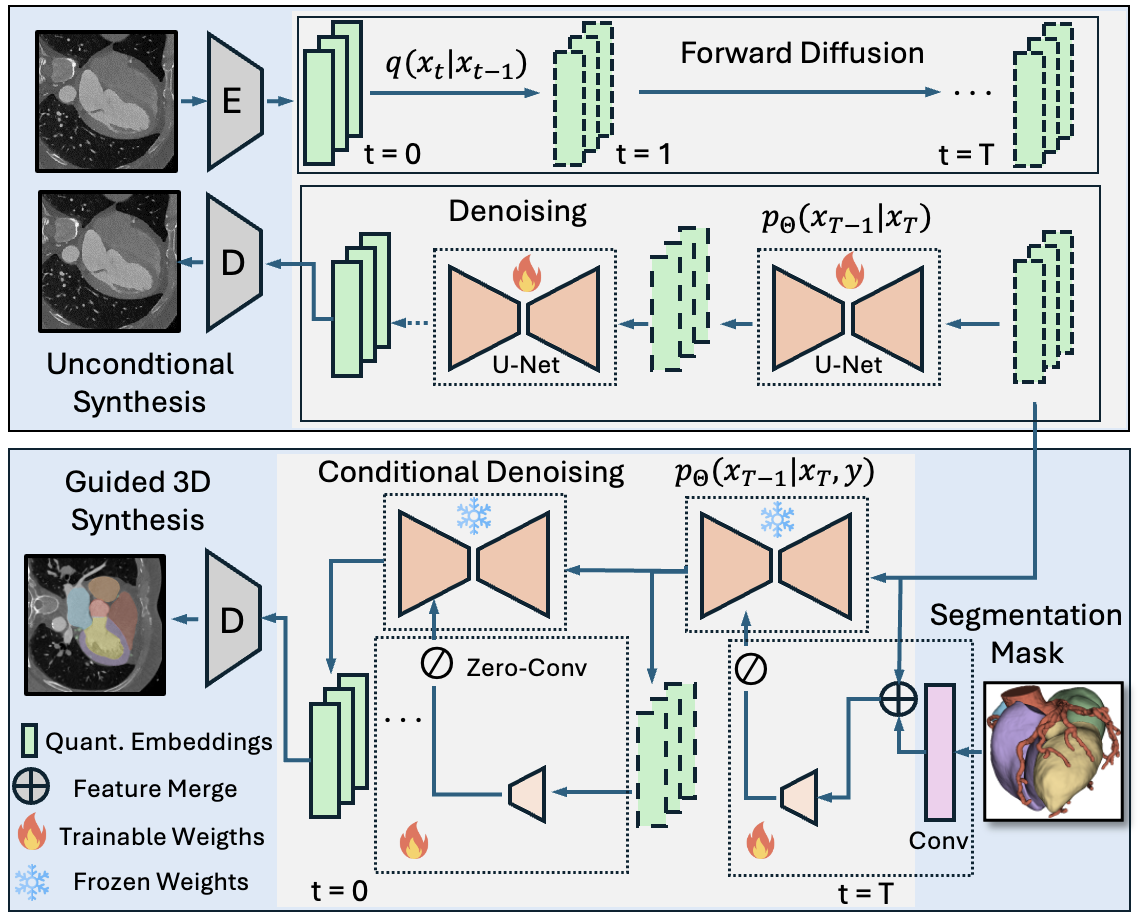}  
    \caption{Top: A 3D U-Net is pre-trained unconditionally in the latent space of an autoencoder. Bottom: A lightweight ControlNet (50\% reduced in size) is trained from scratch to encode external conditions and inject them into the frozen pretrained U-Net via feature modulation at intermediate layers.}
 \label{fig:medlord_architecture}  
\end{figure}
\subsection{Datasets}\label{subsec:Datasets}
\textbf{Photon Counting Coronary Computed Tomography Angiography (PCCTA):} This dataset consisted of 136 CCTA volumes (3D) acquired at the University Medical Centre Mannheim, with ethics approval granted by the Ethics Committee of Ethikkommission II at Heidelberg University (ID 2021-659). 
Among these, 120 volumes were reserved for training and 16 for testing. All volumes were center-cropped to a size of 512$\times$512$\times$256, and  intensity-clipped between -1000 and 2000.
\textbf{Lung CT (LUNA):} This dataset comprised of 888 lung volumes (3D) from the publicly available Luna16 dataset \cite{luna}. Among these, 800 volumes were used for training, and 88 volumes for testing. The volumes were resized to 512$\times$512$\times$256, and intensity-clipped between -1000 and 2000.
\subsection{Baseline Methods}\label{subsec:methods}
We compare MedLoRD to several state-of-the-art generative models, including MAISI \cite{maisi}, HA-GAN \cite{hagan}, and VQ-Transformer \cite{vqvae2transformer}.
To ensure a fair comparison, all baseline methods were adapted to run within a 24 GB GPU memory constraint. This was achieved by either resizing the input data (MAISI, HA-GAN) or increasing the compression ratio of the autoencoder (VQ-Trans).\\
\textbf{MAISI:}
MAISI is built on latent diffusion models, utilizing a variational autoencoder GAN to encode data . The diffusion model is also conditioned on the voxel spacing of the volumes. During synthesis, to manage computational constraints, decoding is performed on 3D patches in the latent space, followed by volume-stitching to reconstruct complete volumes. Training procedures, hyperparameters, and pre-trained foundation autoencoder were adapted from \footnote{\url{https://github.com/Project-MONAI/tutorials/tree/main/generation/maisi}}. For model training, all data samples were downsampled to 512$\times$512$\times$128 to fit within the available GPU VRAM, as measured GPU memory consumption did not align with the reported values in the repository.
Two variants of MAISI were considered. In the first variant, MAISI\textsubscript{ST}, the encoding model was trained from scratch, and in the second variant, MAISI\textsubscript{PT}, a pre-trained foundation encoding model trained on around 40k CT images was adapted. In the PCCTA dataset, MedLoRD was compared to both MAISI\textsubscript{ST} and MAISI\textsubscript{PT}, whereas in the LUNA dataset, only the better-performing MAISI variant (MAISI\textsubscript{PT}) was used due to extensive training times.\\
\textbf{VQ-Trans:} VQ-Trans combines VQ-VAE GAN and transformers. First, a VQ-VAE GAN is trained to encode the image into a latent space. Then, a transformer is trained to sequentially predict voxels in the latent space, starting from the first randomly initialized voxel. Upon sampling in the latent space, a decoder is used to decode the image back into the pixel space. The encoding model was utilized to down-sample the data by 16 to have a sequence length of 12544.\\
\textbf{HA-GAN:} HA-GAN is a GAN-based model that integrates adversarial loss with an additional reconstruction loss for improved performance. All training procedures and hyperparameters were adapted from \footnote{\url{https://github.com/batmanlab/HA-GAN}}.
\section{Results}
\subsection{Unconditional Synthesis}
\begin{figure}[t]  
    \centering
    \includegraphics[width=1\textwidth]{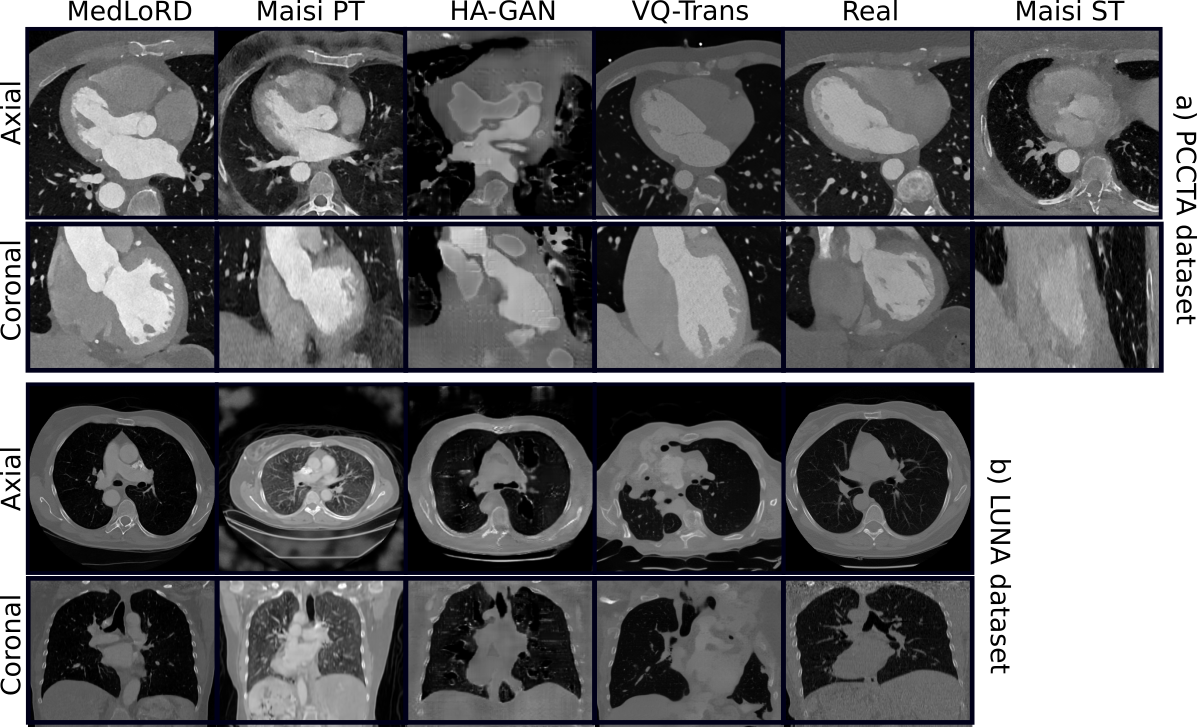}  
    \caption{Randomly selected unconditional samples for each method. Additional MedLoRD samples are provided in the supplementary material.}
 \label{fig:uncond_samples}  
\end{figure}

First, we conducted unconditional image synthesis. Fig. \ref{fig:uncond_samples} shows representative samples from both the PCCTA and LUNA datasets. MedLoRD consistently generates high-quality, realistic images, preserving both global and local structural details. In contrast, other models exhibited unstable behavior, with some samples showing poor quality and heavy artifacts (e.g., loss of heart structures in MAISI\textsubscript{ST}, poor realism in HA-GAN; see Fig. \ref{fig:uncond_pccta}). VQ-Trans and MAISI\textsubscript{PT} showed partial success, though the latter still displayed artifacts in PCCTA samples.

\begin{figure}[t]  
    \centering
    \includegraphics[width=1\textwidth]{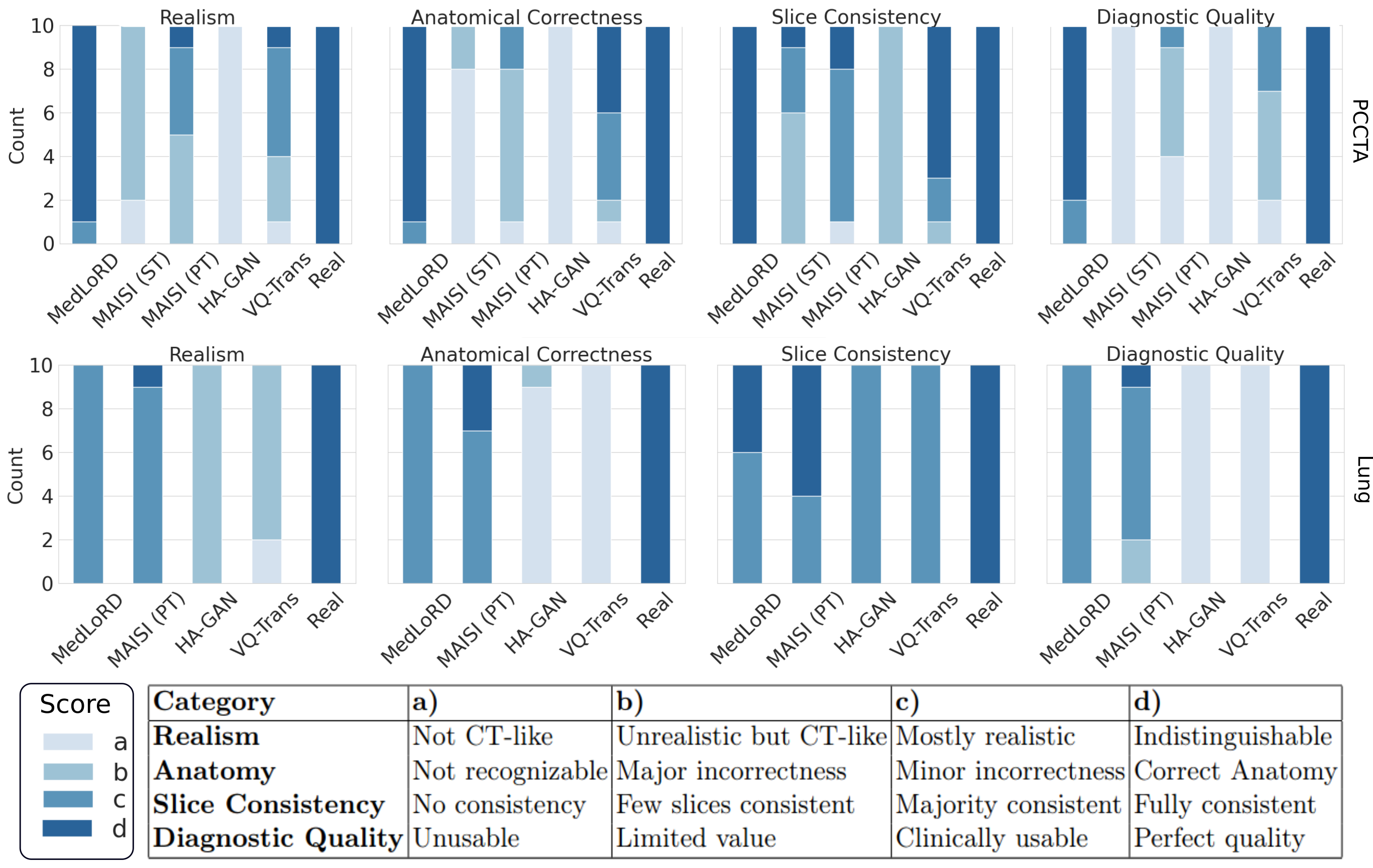}  
    \caption{Radiological evaluation. Top: PCCTA dataset. Bottom: Lung dataset. Images were randomly selected and then shuffled across methods ensuring blinded assessment.}
    \label{fig:radiological}  
\end{figure}

\begin{table}[b]
\centering
\caption{FID Scores for Different Methods on Luna and PCCTA Datasets}
\begin{tabular}{|l|c|c|c|c|c|c|}
\hline
\textbf{Dataset} & \textbf{MedLoRD} & \textbf{MAISI\textsubscript{ST} }& \textbf{MAISI\textsubscript{PT}} &\textbf{HA-GAN}  & \textbf{VQ-Trans} & \textbf{Real} \\ \hline
\textbf{PCCTA}   & \textbf{1.50} & 16.76 & 5.40 & 9.01 & 10.67   & 0.35  \\ 
\textbf{Luna}    & 3.62     & - & \textbf{3.33}          & 5.07         & 5.87             & 0.28  \\ 
\hline
\end{tabular}
\label{tab:fidscores}
\end{table}
To quantitatively assess the quality of the generated images, we computed the FID between synthetic and real samples across different methods (Tab. ~\ref{tab:fidscores}). A 2.5D FID approach was employed, using a feature extraction model pretrained on RadImageNet \cite{radimagenet}. MedLoRD achieved the best score on PCCTA (FID 1.50), and a competitive score on LUNA (FID 3.62), closely trailing MAISI\textsubscript{PT} (FID 3.33). These results indicate that our method produces highly realistic samples, particularly on structurally complex datasets such as PCCTA.
\begin{figure}[t]  
    \centering
    \includegraphics[width=1\textwidth]{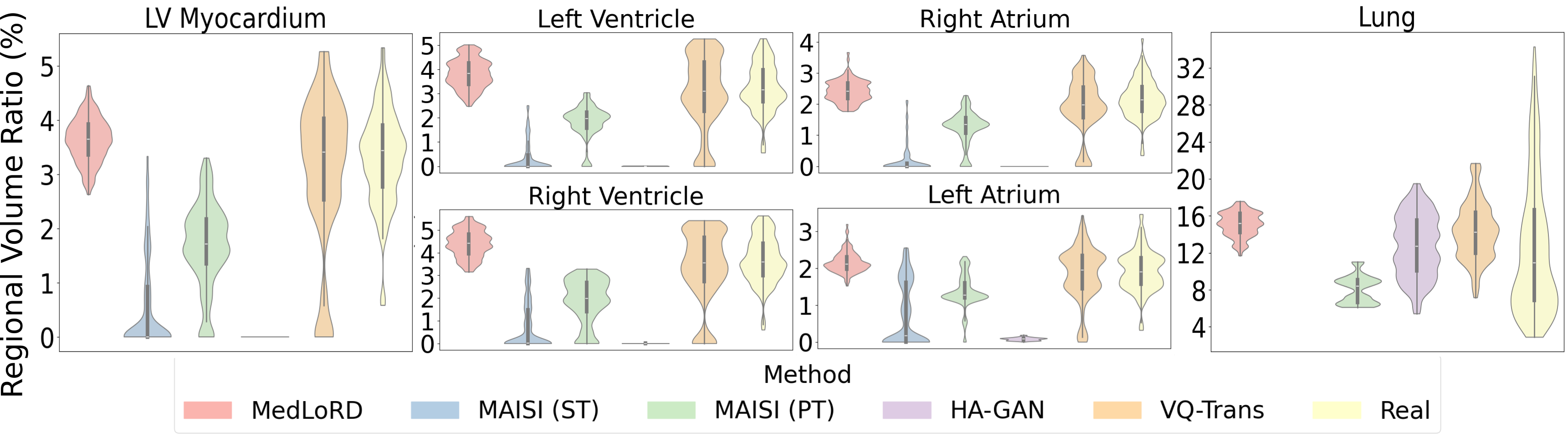}  
    \caption{Regional volume ratio distributions for the PCCTA dataset (left ventricle myocardium, aorta, ventricles, atria, pulmonary artery) and lung dataset.}
    \label{fig:vol_gray}  
\end{figure}
To assess anatomical realism, we included an additional quality metric, the Regional Volume Ratio (RVR). The RVR is calculated by dividing the number of voxels within the masked region of interest by the total number of voxels, assessing whether the generated volumes of regions align with real volume distributions. Segmentation masks for various regions in the PCCTA and LUNA datasets were obtained using TotalSegmentator \cite{totalsegmentatorct}.
These values were then compared to the corresponding values from real images, as provided by the training samples for each dataset. Fig.\ref{fig:vol_gray}  illustrates these comparisons, showing that MedLoRD-generated images exhibited structural distributions closely aligned with real samples. A key observation is that while the distributions in MedLoRD-generated images fall within the real range, they exhibit lower variance. This may be attributed to the use of L1 loss during training, which tends to bias outputs toward the median of the target distribution, thus suppressing variation.
Importantly, we emphasize that evaluation should consider all anatomical structures collectively, rather than isolated regions. A model that consistently generates realistic volumes across multiple structures is more valuable than one that excels in a single region while failing elsewhere (RVR distribution tails close to zero as in HA-GAN, VQ-Trans and MAISI\textsubscript{ST}, indicating that generated samples occasionally lack this anatomical region). MedLoRD demonstrates strong overall consistency, indicating robust anatomical fidelity across diverse structures.\\
Finally, expert radiological assessment (Fig.~\ref{fig:radiological}) confirmed MedLoRD's realism. Each dataset was evaluated by a domain expert following the criteria outlined in Fig.~\ref{fig:radiological}. For each dataset, 10 samples per model (including real data) were randomly selected, resulting in 60 shuffled samples that were anonymized and shown in random order to the radiologist, who graded all evaluation categories per sample. In the PCCTA dataset, MedLoRD outperforms all competing methods, with 8/10 synthesized samples labeled as indistinguishable from real ones, showcasing its strong generative capabilities. In the LUNA dataset, MedLoRD delivers strong performance, with only MAISI\textsubscript{PT} showing slightly higher radiological interest scores. However, 96\% of the 40k CT images used to pre-train MAISI\textsubscript{PT}'s encoding model were chest scans, potentially giving it an advantage in lung image synthesis, due to a more comprehensive latent space. 
Together, these results highlight MedLoRD’s ability to generate anatomically accurate, high-quality 3D medical images, validated by both quantitative and expert evaluations.

\subsection{Conditional Synthesis}
We evaluated MedLoRD’s conditional synthesis by generating volumes from segmentation masks taken from the held-out test set and therefore were never seen during the training of either the unconditional or conditional models.  The generated outputs were segmented using TotalSegmentator, and the resulting masks were compared to the inputs. As shown in Tab.~\ref{tab:cond_metrics}, high DSC scores across regions in both PCCTA and LUNA confirm MedLoRD’s ability to generate anatomically consistent volumes.
Notably, regions with contrast agents (e.g., aorta, LA, LV) showed higher DSC scores, likely due to clearer boundaries and better segmentation accuracy.  Additionally, no data augmentation was applied, which may explain the observed performance gap, as augmentation could potentially improve generalization, especially in smaller datasets like PCCTA.
Fig. \ref{fig:cond_samples} also shows some representative samples adhering to their input segmentation masks.\\
Finally, we evaluated the model on downstream segmentation tasks using nnUNet \cite{nnunet}:
We compared models trained on all real data versus those trained on an equal number of synthetic data.
As shown in Tab. \ref{tab:cond_metrics} performance differences were small ($\leq 0.05$ median DSC), suggesting that synthetic data can serve as a viable surrogate. 

\begin{table}[h]
\centering
\caption{Medians and IQRs of DSC for different input conditions and downstream task (DT) performance on held-out test samples.}
\resizebox{\textwidth}{!}{%
\begin{tabular}{|l|c|c|c|c|c|c|c|}
\hline
\textbf{Metric} & \textbf{Aorta} & \textbf{LA} & \textbf{RA} & \textbf{LV Myo} & \textbf{LV} & \textbf{RV} & \textbf{Lung} \\ \hline
\textbf{DSC} & $0.87 (0.09)$ & $0.85 (0.11)$ & $0.81 (0.09)$ & $0.73 (0.15)$ & $0.89 (0.13)$ & $0.80 (0.08)$ & $0.98 (0.00)$ \\ \hline
\textbf{DSC DT real} & $0.97(0.04)$ & $0.98(0.01)$ & $0.97(0.01)$ & $0.97 (0.02)$ & $0.98 (0.01)$ & $0.98(0.02)$ & $0.99 (0.00)$ \\ 
\textbf{DSC DT syn} & $0.95(0.05)$ & $0.95(0.02)$ & $0.94(0.04)$ & $0.92 (0.03)$ & $0.95 (0.03)$ & $0.94(0.04)$ & $0.98 (0.01)$ \\ 
\hline
\end{tabular}%
}
\label{tab:cond_metrics}
\end{table}

\section{Discussion}
We introduced MedLoRD, a low-resource diffusion model trained on $512 \times 512 \times 256$ volumes, capable of synthesizing high-resolution 3D images on standard 24GB VRAM GPUs.
Extensive evaluations on the PCCTA dataset demonstrated outstanding performance in traditional metrics, as well as generating structures according to real volume distributions. Additionally, our model excelled in radiological evaluation, even when compared to a state-of-the-art generative model with a foundational encoder, MAISI\textsubscript{PT}. To further assess generalizability, we evaluated MedLoRD on the LUNA lung dataset. While the MAISI\textsubscript{PT} model received slightly higher radiological preference, MedLoRD achieved a higher average diagnostic quality score and outperformed all other baseline methods across quantitative metrics. It’s important to note that MAISI\textsubscript{PT} was developed using substantially greater data and computational resources. In contrast, MedLoRD was trained under constrained conditions.
This highlights our core contribution: enabling high-quality, controllable 3D medical image synthesis with limited data and compute. MedLoRD represents a promising tool for both research and clinical data augmentation.
Crucially, we address a key gap in the field—rigorous evaluation of synthetic image quality—by introducing a more comprehensive framework combining multiple metrics and radiological assessments. Our results show the limitations of relying on single metrics, motivating future integration of signals like data memorization \cite{memorization} into our training pipeline.
Finally, conditional generation experiments revealed that substituting real data with MedLoRD-generated samples had minimal impact on downstream tasks, reinforcing the value of our synthetic data. While no statistical test was applied, this challenging setup was intended as a proof of concept to assess conditional synthesis in isolation. In practice, synthetic data is typically used to augment real datasets, and potential distributional shifts may explain minor performance gaps.
\subsubsection{\ackname}

This work was supported through state funds approved by the State Parliament of Baden-Württemberg for the Innovation Campus Health + Life Science Alliance Heidelberg Mannheim, Heidelberg Faculty of Medicine at Heidelberg University, BMBF-SWAG Project 01KD2215D, and by the Multi-DimensionAI project of the Carl Zeiss Foundation (P2022-08-010). The authors acknowledge \textbf{1-} the data storage service SDS@hd supported by the Ministry of Science, Research and the Arts Baden-Württemberg (MWK) and the German Research Foundation (DFG) through grant INST 35/1314-1 FUGG and INST 35/1503-1 FUGG, \textbf{2-} the state of Baden-Württemberg through bwHPC and the German Research Foundation (DFG) through grant INST 35/1597-1 FUGG.  

\clearpage
\appendix
\setcounter{figure}{0}
\renewcommand{\thefigure}{A\arabic{figure}}
\pagestyle{empty}
\section{Appendix}
\subsection{Training Details}
MedLoRD's autoencoders were trained on patches of size 128×128×128 for the PCCTA dataset and 128×128×64 for the LUNA dataset, cropped at randomly sampled centers of the original images, for 250,000 iterations. The number of training iterations was adopted from the respective repository if reported; otherwise, generative models were trained for 750,000 iterations, with the best-performing epoch chosen based on the lowest FID score and visual evaluation. For each evaluation interval, we synthesized 120 samples for PCCTA and 88 samples for LUNA. In MedLoRD, a cosine noise schedule \cite{cosinesched} with T=300 timesteps was used for training diffusion models, and L1-loss was minimized using the AdamW optimizer \cite{adamw}. All experiments, including model training, inference, and evaluation, were designed to be conducted on a single RTX 3090 24 GB GPU with CUDA 12.2.

\subsection{Autoencoder metrics}

The best-performing epoch was selected based on MS-SSIM (with kernel size 3), PSNR, and perceptual LPIPS loss values calculated on the held-out test samples. Results are listed in table \ref{tab:autoencoder_scores}.
\begin{table}[h]
\centering
\caption{Autoencoder evaluation metrics: PSNR (↑), MS-SSIM (↑), LPIPS (↓)}
\begin{tabular}{|l|l|c|c|c|c|}
\hline
\textbf{Dataset} & \textbf{Metric} & \textbf{MedLoRD} & \textbf{MAISI\textsubscript{ST}} & \textbf{MAISI\textsubscript{PT}} &  \textbf{VQ-Trans} \\
\hline
\multirow{3}{*}{PCCTA} 
  & PSNR     & \textbf{44.460(1.322)} & 29.613(1.998) & 33.36(1.050)  & 29.011(1.120) \\
  & MS-SSIM  & \textbf{0.998(0.001)} & 0.963(0.010) & 0.973(0.075)  & 0.896(0.012) \\
  & LPIPS    & \textbf{0.043(0.008)} & 0.099(0.022) & 0.087(0.012)  & 0.325(0.020) \\
\hline
\multirow{3}{*}{Luna} 
  & PSNR     &\textbf{ 40.634(7.220)} & - & 34.631(6.440)  & 30.684(2.361) \\
  & MS-SSIM  & \textbf{0.995(0.010)} & - & 0.982(0.026)  & 0.934(0.022) \\
  & LPIPS    & 0.113(0.171) & - & \textbf{0.077(0.059)}  & 0.343(0.164) \\
\hline
\end{tabular}
\label{tab:autoencoder_scores}
\end{table}

\subsection{Qualitative Results}
\begin{figure}[h] 
    \centering
    \includegraphics[width=0.9\textwidth]{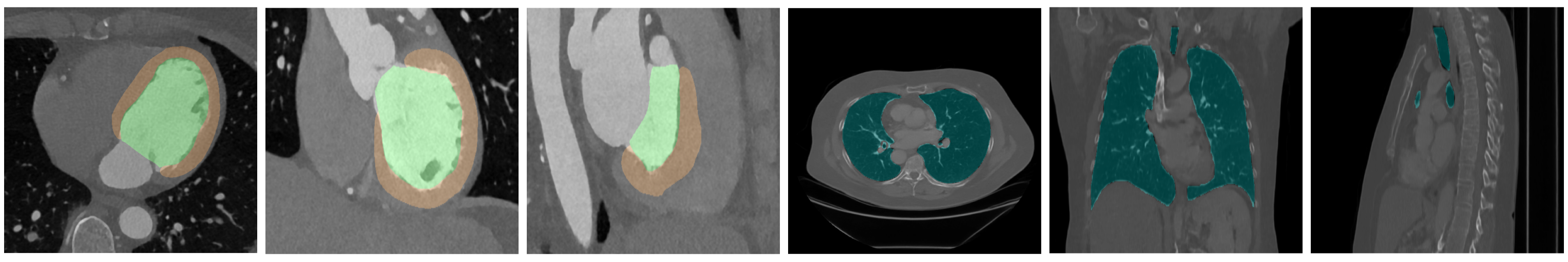}  
    \caption{Conditional samples. Highlighted in color are input condition masks.}
    \label{fig:cond_samples}  
\end{figure}
\begin{figure}[h]
    \centering

    \begin{subfigure}[h]{0.9\textwidth}
        \centering
        \includegraphics[width=\textwidth]{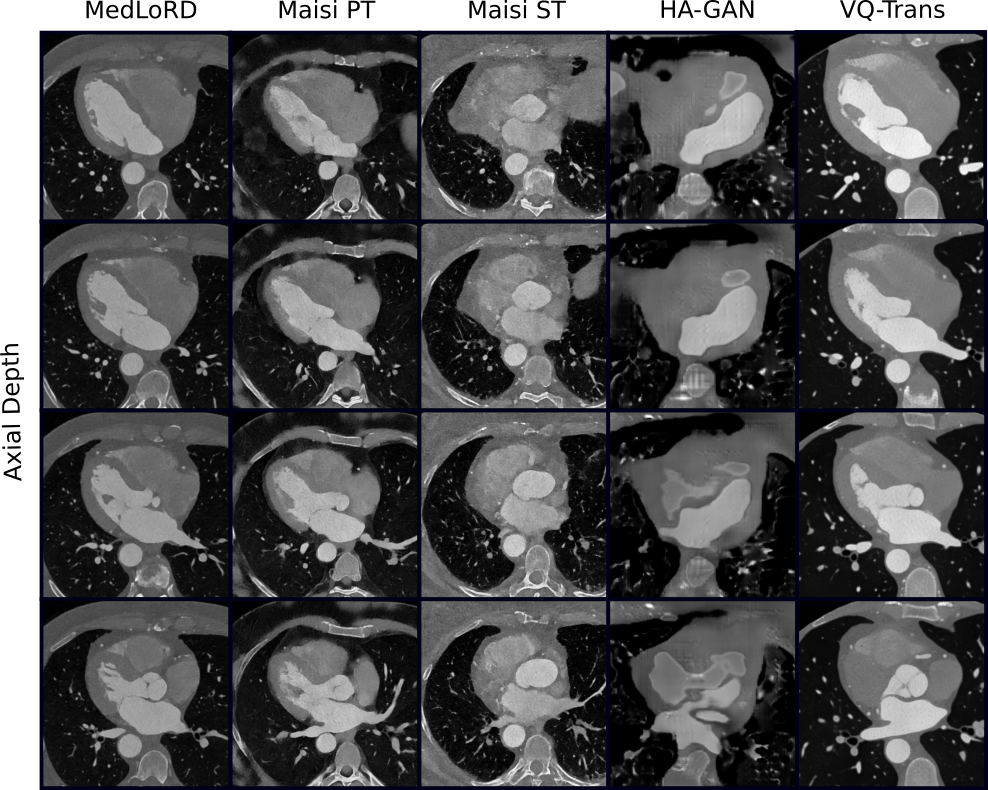}
        \caption{Randomly selected unconditional PCCTA samples (axial view).}
        \label{fig:uncond_pccta}
    \end{subfigure}

    \vspace{0.5em}  

    \begin{subfigure}[h]{0.9\textwidth}
        \centering
        \includegraphics[width=\textwidth]{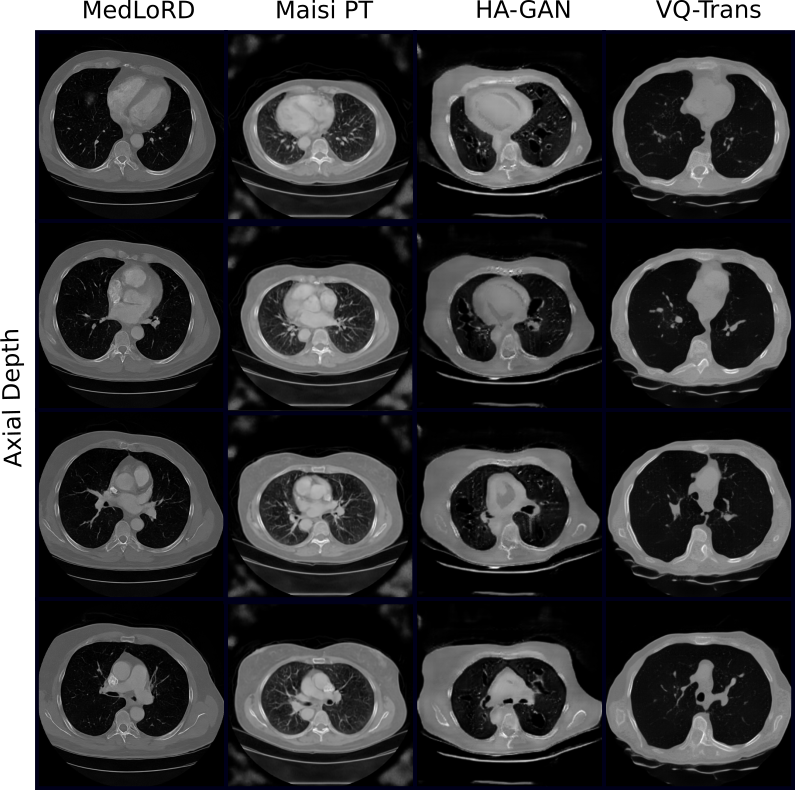}
        \caption{Randomly selected unconditional Luna samples (axial view).}
        \label{fig:uncond_luna}
    \end{subfigure}

    \caption{Randomly selected unconditional samples from PCCTA and Luna datasets.}
    \label{fig:uncond_combined}
\end{figure}

%
%
\clearpage
\FloatBarrier 
\bibliographystyle{splncs04}
\bibliography{bibiliography}

\begin{thebibliography}{10}
\providecommand{\url}[1]{\texttt{#1}}
\providecommand{\urlprefix}{URL }
\providecommand{\doi}[1]{https://doi.org/#1}

\bibitem{monai}
Cardoso, M.J., Li, W., Brown, R., Ma, N., Kerfoot, E., Wang, Y., Murrey, B.,
  Myronenko, A., Zhao, C., Yang, D., Nath, V., He, Y., Xu, Z., Hatamizadeh, A.,
  Myronenko, A., Zhu, W., Liu, Y., Zheng, M., Tang, Y., Yang, I., Zephyr, M.,
  Hashemian, B., Alle, S., Darestani, M.Z., Budd, C., Modat, M., Vercauteren,
  T., Wang, G., Li, Y., Hu, Y., Fu, Y., Gorman, B., Johnson, H., Genereaux, B.,
  Erdal, B.S., Gupta, V., Diaz-Pinto, A., Dourson, A., Maier-Hein, L., Jaeger,
  P.F., Baumgartner, M., Kalpathy-Cramer, J., Flores, M., Kirby, J., Cooper,
  L.A.D., Roth, H.R., Xu, D., Bericat, D., Floca, R., Zhou, S.K., Shuaib, H.,
  Farahani, K., Maier-Hein, K.H., Aylward, S., Dogra, P., Ourselin, S., Feng,
  A.: Monai: An open-source framework for deep learning in healthcare (2022),
  \url{https://arxiv.org/abs/2211.02701}

\bibitem{memorization}
Dar, S.U.H., Seyfarth, M., Ayx, I., Papavassiliu, T., Schoenberg, S.O.,
  Siepmann, R.M., Laqua, F.C., Kahmann, J., Frey, N., Baeßler, B., Foersch,
  S., Truhn, D., Kather, J.N., Engelhardt, S.: Unconditional latent diffusion
  models memorize patient imaging data: Implications for openly sharing
  synthetic data (2025), \url{https://arxiv.org/abs/2402.01054}

\bibitem{dorjsembe2023conditional}
Dorjsembe, Z., Pao, H.K., Odonchimed, S., Xiao, F.: Conditional diffusion
  models for semantic 3d brain mri synthesis. IEEE Journal of Biomedical and
  Health Informatics  (2024). \doi{10.1109/JBHI.2024.3385504}

\bibitem{vqvae2transformer}
Graham, M.S., Tudosiu, P.D., Wright, P., Pinaya, W.H.L., Jean-Marie, U., Mah,
  Y.H., Teo, J.T., Jager, R., Werring, D., Nachev, P., et~al.:
  Transformer-based out-of-distribution detection for clinically safe
  segmentation. In: International Conference on Medical Imaging with Deep
  Learning. pp. 457--476. PMLR (2022)

\bibitem{maisi}
Guo, P., Zhao, C., Yang, D., Xu, Z., Nath, V., Tang, Y., Simon, B., Belue, M.,
  Harmon, S., Turkbey, B., Xu, D.: Maisi: Medical ai for synthetic imaging. In:
  Proceedings of the Winter Conference on Applications of Computer Vision
  (WACV). pp. 4430--4441 (February 2025)

\bibitem{nnunet}
Isensee, F., Jaeger, P.F., Kohl, S.A., Petersen, J., Maier-Hein, K.H.: nnu-net:
  a self-configuring method for deep learning-based biomedical image
  segmentation. Nature methods  \textbf{18}(2),  203--211 (2021)

\bibitem{meddiff}
Khader, F., M{\"{u}}ller-Franzes, G., {Tayebi Arasteh}, S., Han, T.,
  Haarburger, C., Schulze-Hagen, M., Schad, P., Engelhardt, S., Bae{\ss}ler,
  B., Foersch, S., Stegmaier, J., Kuhl, C., Nebelung, S., Kather, J.N., Truhn,
  D.: {Denoising diffusion probabilistic models for 3D medical image
  generation}. Scientific Reports  \textbf{13}(1), ~7303 (2023).
  \doi{10.1038/s41598-023-34341-2}

\bibitem{adamw}
Loshchilov, I., Hutter, F.: Decoupled weight decay regularization. In:
  International Conference on Learning Representations (2019),
  \url{https://openreview.net/forum?id=Bkg6RiCqY7}

\bibitem{radimagenet}
Mei, X., Liu, Z., Robson, P.M., Marinelli, B., Huang, M., Doshi, A., Jacobi,
  A., Cao, C., Link, K.E., Yang, T., Wang, Y., Greenspan, H., Deyer, T., Fayad,
  Z.A., Yang, Y.: Radimagenet: An open radiologic deep learning research
  dataset for effective transfer learning. Radiology: Artificial Intelligence
  \textbf{0}(ja),  e210315 (0). \doi{10.1148/ryai.210315},
  \url{https://doi.org/10.1148/ryai.210315}

\bibitem{cosinesched}
Nichol, A.Q., Dhariwal, P.: Improved denoising diffusion probabilistic models.
  In: Meila, M., Zhang, T. (eds.) Proceedings of the 38th International
  Conference on Machine Learning. Proceedings of Machine Learning Research,
  vol.~139, pp. 8162--8171. PMLR (18--24 Jul 2021),
  \url{https://proceedings.mlr.press/v139/nichol21a.html}

\bibitem{peng2023}
Peng, W., Adeli, E., Bosschieter, T., Park, S.H., Zhao, Q., Pohl, K.M.:
  Generating realistic brain mris via a conditional diffusion probabilistic
  model. In: Medical Image Computing and Computer Assisted Intervention --
  MICCAI 2023. p. 14–24. Springer-Verlag, Berlin, Heidelberg (2023).
  \doi{10.1007/978-3-031-43993-3_2},
  \url{https://doi.org/10.1007/978-3-031-43993-3_2}

\bibitem{flash_attention}
Rabe, M.N., Staats, C.: Self-attention does not need $o(n^2)$ memory (2022),
  \url{https://arxiv.org/abs/2112.05682}

\bibitem{LDM}
Rombach, R., Blattmann, A., Lorenz, D., Esser, P., Ommer, B.: High-resolution
  image synthesis with latent diffusion models (2021)

\bibitem{luna}
Setio, A.A.A., Traverso, A., {de Bel}, T., Berens, M.S., van~den Bogaard, C.,
  Cerello, P., Chen, H., Dou, Q., Fantacci, M.E., Geurts, B., van~der Gugten,
  R., Heng, P.A., Jansen, B., {de Kaste}, M.M., Kotov, V., Lin, J.Y.H.,
  Manders, J.T., Sóñora-Mengana, A., García-Naranjo, J.C., Papavasileiou,
  E., Prokop, M., Saletta, M., Schaefer-Prokop, C.M., Scholten, E.T., Scholten,
  L., Snoeren, M.M., Torres, E.L., Vandemeulebroucke, J., Walasek, N., Zuidhof,
  G.C., van Ginneken, B., Jacobs, C.: Validation, comparison, and combination
  of algorithms for automatic detection of pulmonary nodules in computed
  tomography images: The luna16 challenge. Medical Image Analysis  \textbf{42},
   1--13 (2017). \doi{https://doi.org/10.1016/j.media.2017.06.015},
  \url{https://www.sciencedirect.com/science/article/pii/S1361841517301020}

\bibitem{hagan}
Sun, L., Chen, J., Xu, Y., Gong, M., Yu, K., Batmanghelich, K.: Hierarchical
  amortized gan for 3d high resolution medical image synthesis. IEEE Journal of
  Biomedical and Health Informatics  \textbf{26}(8),  3966--3975 (2022).
  \doi{10.1109/JBHI.2022.3172976}

\bibitem{brainsynth}
Tudosiu, P.D., Pinaya, W.H., Ferreira Da~Costa, P., Dafflon, J., Patel, A.,
  Borges, P., Fernandez, V., Graham, M.S., Gray, R.J., Nachev, P., et~al.:
  Realistic morphology-preserving generative modelling of the brain. Nature
  Machine Intelligence pp.~1--9 (2024). \doi{10.1038/s42256-024-00864-0}

\bibitem{totalsegmentatorct}
Wasserthal, J., Breit, H.C., Meyer, M.T., Pradella, M., Hinck, D., Sauter,
  A.W., Heye, T., Boll, D.T., Cyriac, J., Yang, S., Bach, M., Segeroth, M.:
  Totalsegmentator: Robust segmentation of 104 anatomic structures in ct
  images. Radiology: Artificial Intelligence  \textbf{5}(5),  e230024 (2023).
  \doi{10.1148/ryai.230024}, \url{https://doi.org/10.1148/ryai.230024}

\bibitem{medsyn}
Xu, Y., Sun, L., Peng, W., Jia, S., Morrison, K., Perer, A., Zandifar, A.,
  Visweswaran, S., Eslami, M., Batmanghelich, K.: Medsyn: Text-guided
  anatomy-aware synthesis of high-fidelity 3-d ct images. IEEE Transactions on
  Medical Imaging  \textbf{43}(10),  3648--3660 (2024).
  \doi{10.1109/TMI.2024.3415032}

\bibitem{controlnetxs}
Zavadski, D., Feiden, J.F., Rother, C.: Controlnet-xs: Rethinking the control
  of text-to-image diffusion models as feedback-control systems (2024)

\end{thebibliography}
\end{document}